\documentclass[10pt,conference]{IEEEtran}
\IEEEoverridecommandlockouts
\usepackage{cite}
\usepackage{amsmath,amssymb,amsfonts}

\usepackage{algorithmic}
\usepackage{algorithm} % [新增] 必须要有这个，才能用 \begin{algorithm} 环境

\usepackage{graphicx}
\usepackage{textcomp}
\usepackage{balance}   % [新增] 用于最后一页双栏对齐
\usepackage{url}       % [新增] 优化长 URL 的断行

\usepackage{booktabs}  % [新增] 必须！用于 \toprule, \midrule
\usepackage{multirow}  % [新增] 用于合并行
\usepackage{tabularx}  % [新增] 用于表格宽度的自动调整

\usepackage[table]{xcolor} 

\usepackage[hidelinks]{hyperref} 

\usepackage{tcolorbox}
\tcbuselibrary{skins} % 支持 enhanced 样式
\usepackage{xcolor}
\usepackage{enumitem} % 修复列表缩进

\definecolor{mBlue}{RGB}{105, 129, 168} 
\definecolor{mBlueBack}{RGB}{240, 244, 250} 

\definecolor{mRed}{RGB}{186, 126, 126}
\definecolor{mRedBack}{RGB}{252, 245, 245}

\newtcolorbox{promptbox}[2][]{
  enhanced,
  title={#2},
  colframe=#1,          % 边框颜色 (mBlue 或 mRed)
  colbacktitle=#1,      % 标题背景色 (同边框)
  colback=#1Back,       % 【关键】内容背景色恢复为浅色系
  coltitle=white,       % 标题文字颜色
  fonttitle=\bfseries\small,
  fontupper=\footnotesize, 
  boxrule=0.5mm,        
  arc=1.5mm,            % 圆角
  attach boxed title to top left={yshift=-2mm, xshift=3mm}, 
  boxed title style={boxrule=0pt, arc=1mm, frame code={}},
  equal height group=AB, 
  top=4mm, bottom=2mm, left=2mm, right=2mm,
}

\definecolor{morandiBlue}{RGB}{176, 196, 222} 
\definecolor{morandiGrayGreen}{RGB}{188, 210, 188} 
\definecolor{morandiPink}{RGB}{222, 188, 188} 
\definecolor{morandiBeige}{RGB}{210, 205, 185} 
\definecolor{MorandiOrange}{RGB}{240,210,190}

\def\BibTeX{{\rm B\kern-.05em{\sc i\kern-.025em b}\kern-.08em
    T\kern-.1667em\lower.7ex\hbox{E}\kern-.125emX}}
\begin{document}

\title{ARCE: Augmented RoBERTa with Contextualized Elucidations for NER in Automated Rule Checking
}

\author{\IEEEauthorblockN{Anonymous Authors}}
\author{
    \IEEEauthorblockN{
        Jian Chen\IEEEauthorrefmark{1}\thanks{*Corresponding author.},
        Jiabao Dou\IEEEauthorrefmark{2}
    }
    \IEEEauthorblockA{\IEEEauthorrefmark{1}Ningxia Research Institute of Transport Science, Yinchuan, China}
    \IEEEauthorblockA{\IEEEauthorrefmark{2}Department of Computer Science, Hong Kong Baptist University, Hong Kong}
}

\maketitle

\begin{abstract}
Accurate information extraction from specialized texts is a critical challenge for automated rule checking (ARC) in the architecture, engineering, and construction (AEC) domain. While large language models (LLMs) possess strong reasoning capabilities, their deployment in resource-constrained AEC environments is often impractical. Conversely, standard efficient models struggle with the significant domain gap. Although this gap can be mitigated by pre-training on large, human-curated corpora, such approaches are labor-intensive and costly. To address this, we propose ARCE (Augmented RoBERTa with Contextualized Elucidations), a novel knowledge distillation framework that leverages LLMs to synthesize a task-oriented corpus, termed \textit{Cote}, for incrementally pre-training smaller models. ARCE systematically explores the optimal strategy for knowledge transfer. Our extensive experiments demonstrate that ARCE establishes a new state-of-the-art on a benchmark AEC dataset, achieving a Macro-F1 score of 77.20\% and outperforming both domain-specific baselines and fine-tuned LLMs. Crucially, our study reveals a less is more principle: simple, direct explanations prove significantly more effective for domain adaptation than complex, role-based rationales in the NER task, which tend to introduce semantic noise. The source code will be made publicly available upon acceptance.
\end{abstract}


\begin{figure*}
\centering
\includegraphics[width=0.9\textwidth]{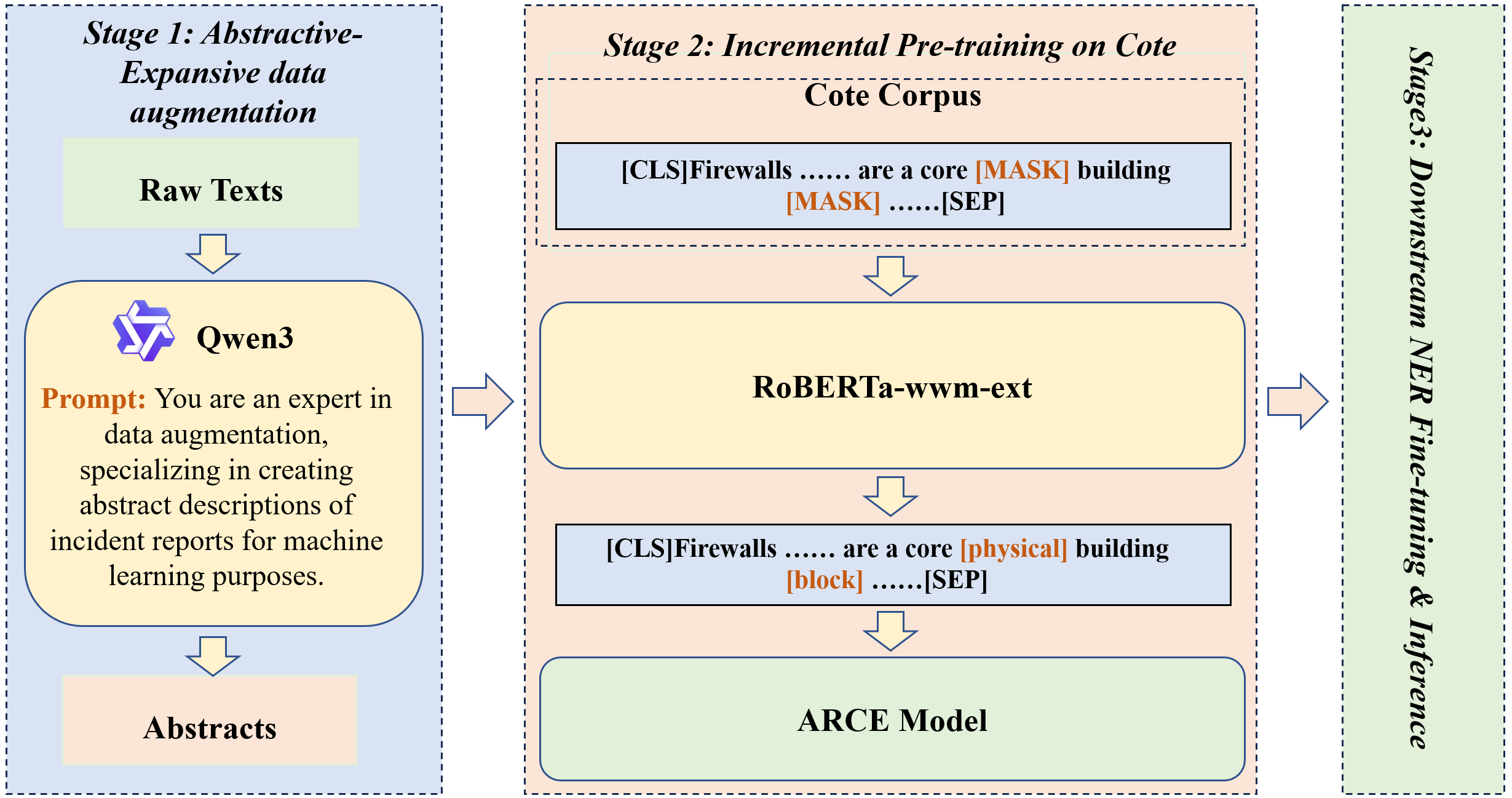}
\caption{The overall architecture of our proposed ARCE approach. The approach consists of three stages. \textbf{Stage 1}: We employ a LLM (e.g., Qwen3) to generate a Cote (contextualized task-oriented elucidation) corpus from raw domain texts using specialized prompts. \textbf{Stage 2}: A pre-trained RoBERTa-wwm-ext model undergoes incremental pre-training on the Cote corpus via a masked language modeling (MLM) objective, yielding the enhanced ARCE model. \textbf{Stage 3}: The ARCE model is augmented with a CRF layer and fine-tuned on the downstream NER task, and is then used for inference on new, unseen texts.}
\end{figure*}

\section{INTRODUCTION}

The architecture, engineering, and construction (AEC) industry, a cornerstone of the global economy, operates under a complex framework of safety codes and regulatory requirements documented in vast volumes of unstructured text~\cite{ismail2017review,zheng2022pretrained,zhong2024domain}. To navigate this complexity and mitigate human error, automated rule checking (ARC) has emerged as a vital technology for ensuring regulatory compliance~\cite{wu2022natural,song2018nlp}. A critical bottleneck in developing effective ARC systems, however, lies in the rule interpretation stage. This stage necessitates the accurate extraction of semantic information from specialized texts---a task formally known as named entity recognition (NER)~\cite{zhou2022integrating,chen2026qwen}.

Early attempts to address the NER challenge in the AEC domain primarily involved fine-tuning pre-trained language models (PLMs) such as BERT~\cite{devlin2019bert} and RoBERTa~\cite{liu2019roberta} on small, in-domain labeled datasets. While these models have achieved state-of-the-art (SOTA) performance in general domains, their effectiveness in specialized fields like AEC is often constrained by a significant domain gap~\cite{zhang2021deep,gao2025transfer}. The specialized lexicon and complex syntactic structures found in regulation documents differ drastically from the general-purpose corpora used for the models' initial training. This mismatch limits the models' ability to capture domain-specific semantic dependencies, making robust information extraction a persistent challenge.

To bridge this domain gap, the research community has largely converged on two primary paradigms. The first is domain-adaptive pre-training (DAPT), exemplified by approaches like ARCBERT~\cite{zheng2022pretrained}. This method involves further pre-training general PLMs on massive, human-curated domain-specific corpora to inject domain knowledge \cite{gururangan2020dapt}. While effective, DAPT is exceptionally labor-intensive and costly. Collecting and cleaning gigabytes of high-quality AEC texts requires significant domain expertise and engineering effort, making it difficult to scale or update as regulations evolve \cite{zheng2024text}.

The second, more recent paradigm leverages the emergent capabilities of large language models (LLMs). Models such as GPT-5 \cite{achiam2023gpt4} and Qwen3~\cite{qwen3technicalreport} possess strong reasoning and zero-shot extraction capabilities. However, deploying billion-parameter LLMs directly for ARC tasks presents severe practical hurdles. In real-world AEC scenarios---often involving on-site devices or high-throughput document processing---the exorbitant computational cost, high inference latency, and memory footprint of LLMs are prohibitive. 

Consequently, a new research frontier has opened: how to leverage the knowledge of LLMs to augment smaller, more efficient models \cite{hsieh2023distilling}. This is often achieved through generative data augmentation \cite{ye2022zerogen,wang2021want}, where LLMs are used to synthesize labeled data or auxiliary knowledge to train smaller models ~\cite{bogdanov2024nuner,zhang2024cross}. However, a critical scientific question remains unexplored in this context: What is the optimal format of knowledge for transfer? While the community has heavily focused on Chain-of-Thought (CoT) prompting to elicit complex reasoning from LLMs \cite{wei2022chain,guo2025deepseek-r1}, it is unclear whether such verbose and complex rationales are beneficial or detrimental when distilling knowledge into a smaller model with limited capacity for the NER task \cite{turpin2023unfaithful}. 

In this work, we address these challenges by proposing ARCE (Augmented RoBERTa with Contextualized Elucidations). Unlike traditional DAPT methods that rely on raw text, ARCE introduces a novel knowledge distillation framework that leverages an LLM to generate a high-quality, task-oriented knowledge corpus---termed Cote. Instead of simply generating more training labels, ARCE tasks the LLM with providing contextualized elucidations---explicit semantic explanations of why a specific span constitutes an entity within its context. This corpus is then used to incrementally pre-train a RoBERTa model, effectively aligning its representation space with the domain-specific logic before fine-tuning.

Our approach offers a cost-effective alternative to massive pre-training while maintaining the inference efficiency of small language models (SLMs). Through extensive experiments, we validate that ARCE establishes a new state-of-the-art on a benchmark AEC dataset, outperforming both domain-adapted baselines and directly fine-tuned LLMs. 

Crucially, our investigation yields a counter-intuitive and valuable insight regarding knowledge generation. We systematically compare different prompting strategies—ranging from complex, step-by-step reasoning to simple, direct explanations. We discover a principle: for domain adaptation of SLMs, a large volume of simple, explanation-based elucidations is significantly more effective than complex, role-based rationales. We hypothesize that while complex reasoning helps LLMs themselves, it introduces noise and overfitting risks for smaller models, whereas simple elucidations provide robust, linear semantic features that are easier to digest.

The main contributions of this paper are summarized as follows:
\begin{itemize}
    \item We propose ARCE, a novel framework that leverages LLM-generated elucidations to effectively adapt a RoBERTa model to the AEC domain. This approach circumvents the high costs of collecting large human-curated corpora required by methods like ARCBERT and avoids the deployment latency of LLMs.
    \item We conduct a systematic ablation study of knowledge generation strategies. We reveal that simple, explanation-based elucidations significantly outperform complex, role-based reasoning for SLM adaptation, offering a guiding principle for future generative data augmentation research in the NER task.
    \item We establish a new state-of-the-art performance on the benchmark AEC NER dataset, surpassing strong baselines including domain-adapted BERT models and fine-tuned 8B-parameter LLMs, while maintaining a fraction of the inference cost.
\end{itemize}

\section{METHODOLOGY}

Our proposed approach, ARCE (augmented RoBERTa with contextualized elucidations), is designed to enhance NER performance in specialized domains by leveraging knowledge generated from a large language model, specifically Qwen3-8B~\cite{qwen3technicalreport}. As illustrated in Figure 1, the ARCE approach is decomposed into three sequential stages: (1) the generation of a contextualized task-oriented elucidation (Cote) corpus; (2) incremental pre-training of a RoBERTa model on this corpus; and (3) downstream NER fine-tuning with a conditional random field (CRF) layer.

\subsection{Stage 1: Contextualized Task-Oriented Elucidation (Cote) Generation}

The cornerstone of our approach is the creation of a high-quality, task-specific corpus, which we term the Corpus of contextualized task-oriented elucidations ($\mathcal{C}_{Cote}$). Unlike traditional data augmentation methods that rely on generic external corpora, we employ a powerful LLM, denoted as $\mathcal{M}$, to generate elucidations explicitly tailored to the reasoning process of the NER task.

Given a sentence $x = \{w_1, w_2, \dots, w_n\}$ from a source dataset and an entity within it, defined by its text span $s$ and entity type $t$, we construct a specialized prompt $P$ based on a prompting strategy $\Pi$, such that $P = \Pi(x, s, t)$. This strategy directs the LLM to generate a textual elucidation $k$ that clarifies why the span $s$ corresponds to the type $t$ within the context of $x$, or describes the functional role it plays.

The generation of the elucidation text $k = \{k_1, k_2, \dots, k_L\}$ is an auto-regressive process, where the probability of the sequence is conditioned on the prompt $P$ and the LLM's parameters $\theta_{\mathcal{M}}$. This is formally expressed as:
\begin{equation}
    p(k | P; \theta_{\mathcal{M}}) = \prod_{i=1}^{L} p(k_i | k_{<i}, P; \theta_{\mathcal{M}})
\end{equation}
By iterating this process over all entities in our source data, we compile the final Cote corpus, $\mathcal{C}_{Cote}$, which serves as the foundation for adapting our NER model:
\begin{equation}
    \mathcal{C}_{Cote} = \{k_j | k_j \sim p(k | \Pi(x_j, s_j, t_j); \theta_{\mathcal{M}})\}
\end{equation}
where $(x_j, s_j, t_j)$ represents the $j$-th entity instance in the source dataset.

\subsection{Stage 2: Incremental Pre-training on Cote}

To effectively integrate the domain-specific knowledge encapsulated in the elucidations into our NER model, we perform incremental pre-training on a RoBERTa model~\cite{cui2020revisiting}. Let its initial parameters be $\theta_{RoBERTa}$. This stage adapts the general-domain model to the specific vocabulary, syntax, and reasoning patterns present in $\mathcal{C}_{Cote}$.

The training objective for this phase is the standard masked language modeling (MLM) task~\cite{devlin2019bert}. For each elucidation text $k \in \mathcal{C}_{Cote}$, we generate a corrupted version $\tilde{k}$ by randomly masking a subset of its tokens. Let $k_{masked}$ be the set of original tokens at the masked positions. The MLM loss, $\mathcal{L}_{MLM}$, is defined as the negative log-likelihood of correctly predicting these original tokens from the corrupted context $\tilde{k}$:
\begin{equation}
    \mathcal{L}_{MLM}(\theta) = - \sum_{k \in \mathcal{C}_{Cote}} \sum_{w \in k_{masked}} \log p(w | \tilde{k}; \theta)
\end{equation}
The RoBERTa model's parameters are updated by minimizing this loss function over the entire Cote corpus. This procedure yields an enhanced set of parameters, $\theta_{ARCE}$, which now encode both general linguistic understanding and the targeted task-specific knowledge, as shown in Eq. (4).
\begin{equation}
    \theta_{ARCE} = \arg\min_{\theta_{RoBERTa}} \mathcal{L}_{MLM}(\theta_{RoBERTa})
\end{equation}

\subsection{Stage 3: Downstream NER Fine-tuning with CRF}

The final stage involves fine-tuning the adapted ARCE model for the downstream NER task. To effectively model dependencies between adjacent labels, we augment the architecture with a conditional random field (CRF) layer. The CRF layer computes a score $s(x, y)$ for a label sequence $y$ by combining emission scores from the encoder and a learned transition matrix.
\begin{equation}
    s(x, y) = \sum_{i=1}^{n} E_{i, y_i} + \sum_{i=1}^{n-1} A_{y_i, y_{i+1}}
\end{equation}

The model is jointly optimized by minimizing the negative log-likelihood of the ground-truth sequence $y_{\text{true}}$.
\begin{equation}
    p(y | x) = \frac{\exp(s(x, y))}{\sum_{y' \in \mathcal{Y}(x)} \exp(s(x, y'))}
\end{equation}
\begin{equation}
    \mathcal{L}_{\text{NER}} = - \log p(y_{\text{true}} | x)
\end{equation}

During inference, the Viterbi algorithm efficiently finds the label sequence $\hat{y}$ that maximizes this score.
\begin{equation}
    \hat{y} = \underset{y' \in \mathcal{Y}(x)}{\arg\max} \, s(x, y')
\end{equation}

\begin{table*}[htbp]
\centering
\caption{Results comparison with different methods. \textbf{Bolded} values indicate the best performance.}
% \renewcommand{\arraystretch}{1.25}
% \scalebox{0.9}{
% \begin{tabular}{lccccccc}
% --- 关键修改 1：消除 booktabs 产生的白色间隙 ---
\setlength{\aboverulesep}{0pt} 
\setlength{\belowrulesep}{0pt}
\renewcommand{\arraystretch}{1.25} % 建议稍微增加行高，避免文字贴着线太紧

\begin{tabularx}{0.95\textwidth}{@{}Xcccccc@{}}
\toprule
% 这里的 gray!25 是背景色
\rowcolor{gray!25}
\textbf{Models} 
&\multicolumn{3}{c}{\textbf{Macro-F1 with Strict Match (\%)}} 
&\multicolumn{3}{c@{}}{\textbf{Macro-F1 with Partial Match (\%)}} \\ % <--- 注意这里的 c 变成了 c@{}

\cmidrule(r){2-4} \cmidrule(l){5-7}

\rowcolor{gray!25}
& Precision   & Recall   & \textbf{Macro-F1} &  Precision   & Recall   & \textbf{Macro-F1} \\
\midrule
\multicolumn{3}{c}{\textit{Traditional Methods}} \\
NuNER-CRF \cite{bogdanov2024nuner} & 64.61 & 62.79 & 63.69  \\
RoBERTa-CRF \cite{liu2019roberta} & 66.94 & 62.35 & 64.56  & & / &  \\
% RoBERTa-wwm-ext-CRF  & 77.49 & 73.76 & 75.59  & & / & \\
BERT-base-Chinese-CRF \cite{devlin2019bert}  & 77.54 & 71.02 & 72.93 \\
RoBERTa-wwm-ext-CRF \cite{cui2020revisiting}  & 74.62 & 68.36 & 71.19  & &  & \\
\midrule
\multicolumn{3}{c}{\textit{8B Large Language Models}} \\
Llama3.1-8B-Instruct \cite{grattafiori2024llama3}  & 21.71 & 10.65 & 13.00 & 31.93 & 20.31 & 22,91 \\
Ministral-8B-Instruct \cite{jiang2023mistral7b} & 22.31 & 16.50 & 18.87 & 35.18 & 33.61 & 34.11 \\
Qwen3-8B \cite{qwen3technicalreport}& 29.09 & 21.98 & 25.01 & 61.93 & 30.77 & 35.73 \\
\midrule
\multicolumn{3}{c}{\textit{Large Language Models for Reasoning}} \\
DS-R1-Distill-Llama-8B \cite{guo2025deepseek-r1}  & 13.79 & 9.63 & 10.51 & 34.90 & 23.56 & 25.39 \\
DS-R1-Distill-Qwen-7B \cite{guo2025deepseek-r1}  & 19.26 & 10.00 & 12.47 & 31.86 & 17.41 & 21.23  \\
DS-R1-0528-Qwen3-8B \cite{guo2025deepseek-r1} & 25.98 & 11.37 & 15.60 & 42.11 & 15.73 & 21.99 \\
Qwen3-8B-think \cite{qwen3technicalreport} & 41.28 & 28.77 & 32.55 & 55.16 & 36.73 & 41.68 \\
Qwen3-235B-A22B-Instruct-2507 \cite{zheng2024llamafactory} & 29.98 & 28.14 & 28.65 & 71.98 & 46.08 & 49.13 \\
\midrule
Qwen3-8B-SFT \cite{zheng2024llamafactory} & 75.48 & 75.39 & 75.23 & 84.07 & 83.41 & \textbf{83.49} \\
ARCBERT-CRF \cite{zheng2022pretrained} & 77.31 & 75.26 & 75.75 & & / &   \\
\rowcolor{MorandiOrange} ARCE(Ours) & 77.51 & 77.30 & \textbf{77.20}  & & / &   \\
% \rowcolor{MorandiOrange} $Ours_{think}$ & 76.46 & 77.67 & 76.86  & & / &   \\
% \rowcolor{MorandiOrange} $Ours_{role}$ & 78.73 & 72.88 & 75.34  & & / &   \\
% \midrule
% \rowcolor{MorandiOrange} $Ours_{25}$ & 76.64 & 72.39 & 74.23  & & / &   \\
% \rowcolor{MorandiOrange} $Ours_{50}$ & 76.10 & 73.33 & 74.40  & & / &   \\   \rowcolor{MorandiOrange} $Ours_{75}$ & 76.47 & 76.54 & 76.30  & & / &   \\
\bottomrule
\end{tabularx}
% }
\end{table*}

\section{EXPERIMENTS}

\subsection{Experimental Setup}
We evaluate our approach on the public AEC-domain NER dataset introduced by Zhou et al.\footnote{\url{https://github.com/zhouyc98/auto-rule-transform}}~\cite{zhou2022integrating}. This dataset serves as a standard benchmark for regulatory compliance checking, containing 4,336 labeled entities distributed across 611 sentences drawn from diverse construction specifications. The entities cover specialized categories such as building components, constraints, and properties. Following the standard protocol established in previous works~\cite{zheng2022pretrained}, we adopt an 8:1:1 split for training, validation, and testing. We report the Macro-F1 score as the primary metric to rigorously account for the label imbalance inherent in specialized domain corpora.

Our implementation leverages the PyTorch framework and the HuggingFace Transformers library. In the incremental pre-training stage, we train the base \texttt{RoBERTa-wwm-ext} model on our generated \texttt{Cote} corpus for 5 epochs. We utilize a learning rate of \texttt{5e-5}, a weight decay of \texttt{0.01} to prevent overfitting, and a batch size of 16. The masked language modeling (MLM) probability is set to 15\%.

For the downstream NER fine-tuning, we train the augmented ARCE model for 10 epochs using the AdamW optimizer with a linear learning rate scheduler. To stabilize the training of the CRF layer, we employ a differential learning rate strategy: the CRF transition parameters are updated with a larger learning rate of \texttt{5e-1}, while the encoder parameters use a finer learning rate of \texttt{5e-5}. All experiments were conducted on a single NVIDIA GeForce RTX 4090 GPU, demonstrating the computational efficiency of our approach compared to multi-GPU LLM training.

\subsection{Baselines}
To comprehensively assess the effectiveness of ARCE, we benchmark it against three distinct categories of baseline models, covering both traditional discriminative methods and cutting-edge generative paradigms:

\begin{enumerate}
    \item \textbf{Traditional Fine-tuned Encoders:} We include standard pre-trained language models fine-tuned with a CRF layer, including \texttt{BERT-base-Chinese}~\cite{devlin2019bert} and \texttt{RoBERTa-wwm-ext}~\cite{liu2019roberta}. These represent the widely used pre-train then fine-tune paradigm without domain adaptation.
    
    \item \textbf{Large Language Models:} We evaluate both \textit{Zero-Shot} and \textit{Fine-Tuned (SFT)} performance of modern LLMs.
    \begin{itemize}
        \item \textit{General LLMs:} We test \texttt{Qwen3-8B}~\cite{qwen3technicalreport}, \texttt{Llama3.1-8B} \cite{grattafiori2024llama3}, and \texttt{Ministral-8B} \cite{jiang2023mistral7b} in a zero-shot setting to gauge their inherent domain understanding.
        \item \textit{Reasoning Models:} We include the recently proposed \texttt{DeepSeek-R1-Distill} series \cite{guo2025deepseek-r1}, which are optimized for complex reasoning chains, to investigate if heavy reasoning aids NER tasks.
        \item \textit{Fine-Tuned LLM:} We include \texttt{Qwen3-8B-SFT}~\cite{zheng2024llamafactory}, a model fine-tuned on the training set with LoRA \cite{hu2021lora}, serving as a high-resource upper bound for generative approaches.
    \end{itemize}
    
    \item \textbf{Domain-Specific SOTA Methods:} We compare against \texttt{ARCBERT}~\cite{zheng2022pretrained}, which uses a large human-curated domain corpus for pre-training, and \texttt{NuNER}~\cite{bogdanov2024nuner}, a state-of-the-art method employing LLM-annotated data for encoder pre-training.
\end{enumerate}

\begin{figure*}
    \centering
    % === 左侧栏 (占用 49% 宽度) ===
    \begin{minipage}[t]{0.49\textwidth}
        \begin{promptbox}[mBlue]{Strategy A: Explanation-Based}
            \textbf{System Instruction:} \\
            You are a world-class requirements documentation analysis expert. 
            
            \vspace{0.5em}
            \textbf{Task:} \\
            Please explain why `\texttt{\{span\}}' can be labeled as `\texttt{\{entity\_type\}}' in the given text.
            
            \vspace{0.5em}
            \textbf{Constraint:} \\
            Your explanation must be logical, direct, and well-grounded in the context.
            
            \vspace{0.8em}
            % 分割线：颜色加深一点点以适应背景
            {\color{mBlue!40}\hrule height 0.8pt} 
            \vspace{0.5em}
            
            \textit{\textbf{Applied Models:}}
            \begin{itemize}[leftmargin=*, nosep]
                \item \textbf{ARCE:} Uses this prompt directly.
                \item \textbf{ARCE-think:} Activates thinking mode.
            \end{itemize}
        \end{promptbox}
    \end{minipage}% 
    \hfill % 弹性间距
    % === 右侧栏 (占用 49% 宽度) ===
    \begin{minipage}[t]{0.49\textwidth}
        \begin{promptbox}[mRed]{Strategy B: Role-Based}
            \textbf{System Instruction:} \\
            You are a world-class requirements analysis expert. Your task is not simply to judge correctness, but to deeply analyze what \textbf{role} a fragment plays, what \textbf{functional effect} it has, and what \textbf{relationships} it has within the complete ``text''.
            
            \vspace{0.5em}
            \textbf{Reference Definitions:} \\
            - Target Type: \texttt{\{label\_en\}} \\
            - Definition: \texttt{\{label\_def\}}
            
            \vspace{0.5em}
            \textbf{Analysis Task:} \\
            - Full Text: ``\texttt{\{text\}}'' \\
            - Target Fragment: ``\texttt{\{span\}}''
            
            \vspace{0.5em}
            \textbf{Constraint:} \\
            Deeply analyze the \textbf{role}, \textbf{function}, and \textbf{relationships}.
        \end{promptbox}
    \end{minipage}
    
    \caption{Comparison of knowledge generation strategies.
    \textbf{Left (Strategy A):} The prompt used in our ARCE framework, focusing on simple explanations. 
    \textbf{Right (Strategy B):} A complex prompt design forcing deep role analysis.}
    \label{fig:prompt_comparison}
\end{figure*}

\subsection{Main Results and Analysis}
The primary experimental results are presented in Table I. Our proposed ARCE approach establishes a new SOTA on the benchmark dataset, achieving a strict-match Macro-F1 score of 77.20\%. ARCE significantly outperforms traditional methods, surpassing the base \texttt{RoBERTa-wwm-ext-CRF} by a substantial margin of 6.01 points. This improvement directly validates the efficacy of our \texttt{Cote} corpus in bridging the domain gap. Furthermore, ARCE surpasses the previous domain-specific SOTA, \texttt{ARCBERT} (75.26\%), by roughly 2 points. Notably, ARCE achieves this without the need for the massive, human-curated corpora that \texttt{ARCBERT} relies on, proving the high value of synthetic, explanation-based knowledge.

A striking observation from Table I is the poor performance of zero-shot LLMs under the strict match criterion. While generative models possess strong semantic understanding, they struggle with precise boundary detection---often including punctuation or extra adjectives in the extracted span. When evaluated under a partial match metric, \texttt{Qwen3-8B}'s score improves to 35.73\%, and the fine-tuned \texttt{Qwen3-8B-SFT} reaches a high 83.49\%. However, in engineering contexts where strict compliance is required, extraction precision is non-negotiable. ARCE, being a token-level discriminative model, inherently avoids these boundary drift issues, offering a more reliable solution for automated rule checking.

Furthermore, we observe that reasoning models perform surprisingly poorly, scoring even lower than standard LLMs. Inspection of their outputs reveals that these models tend to over-think simple extraction tasks \cite{sui2025overthink}, often outputting verbose rationales mixed with the extraction results. This reinforces our finding that complex reasoning is not always beneficial for low-level structural tasks. Finally, comparing ARCE with \texttt{Qwen3-8B-SFT} (75.75\%) highlights a critical efficiency advantage. While the fine-tuned 8B-parameter LLM achieves competitive results, it requires massive computational resources for inference. ARCE, based on the lightweight RoBERTa architecture, achieves a higher accuracy with less than 2\% of the parameter count, making it uniquely suitable for deployment in resource-constrained AEC environments where running an 8B model is infeasible.

\section{Discussion}

\begin{figure}
\centering
\includegraphics[width=0.45\textwidth]{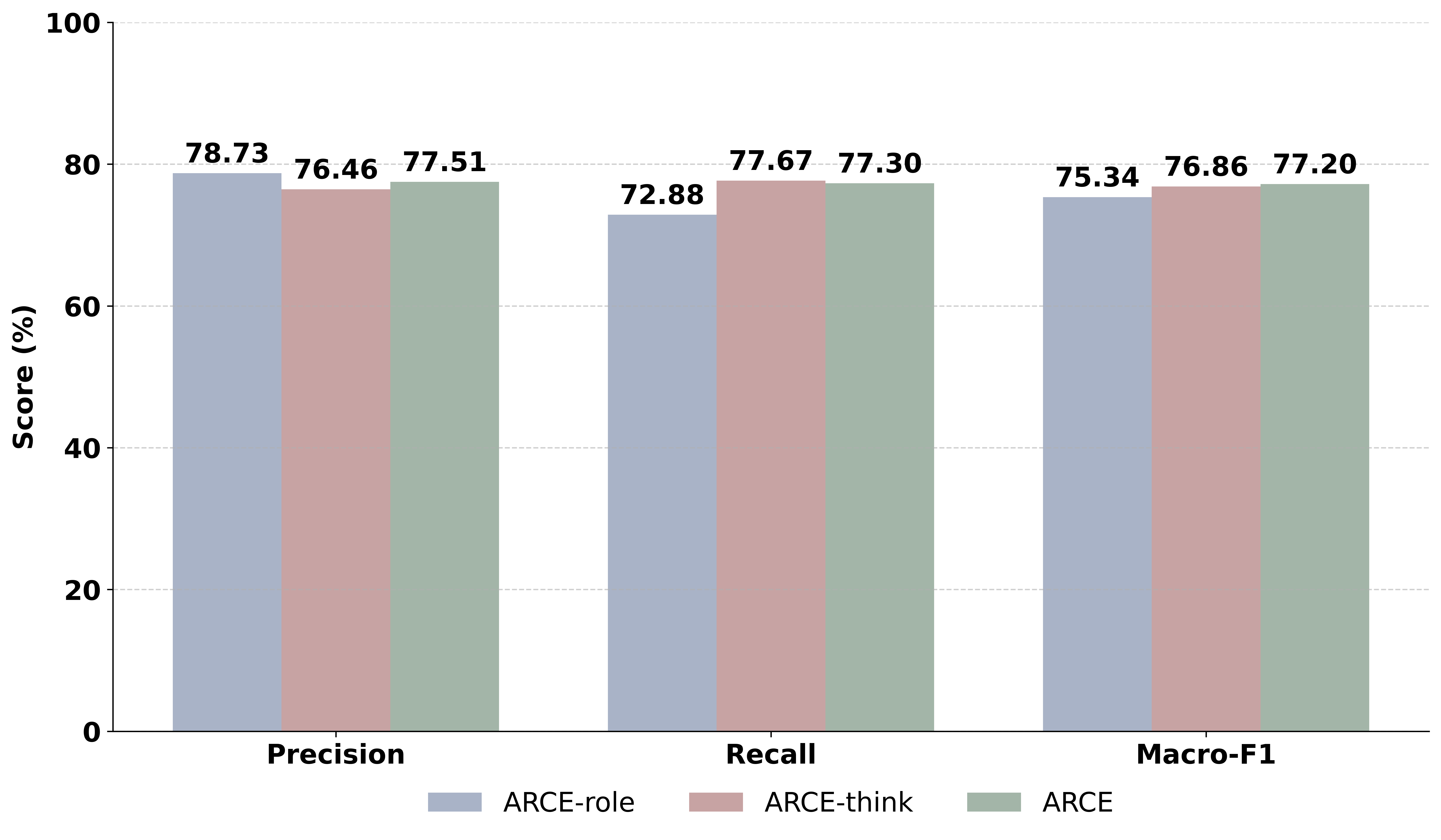}
\caption{Results of the ablation experiment.}
\end{figure}

\subsection{Ablation on Knowledge Generation Strategy}

To isolate the impact of the knowledge generation strategy and validate the less is more hypothesis, we conducted a systematic ablation study comparing our main model, \texttt{ARCE}, with two variants derived from distinct prompting paradigms. \texttt{ARCE} utilizes \textit{Strategy A}, generating simple, direct semantic explanations. The variants include \texttt{ARCE-think}, which adds CoT reasoning traces to Strategy A, and \texttt{ARCE-role}, which employs \textit{Strategy B} to analyze complex functional roles.

As presented in Table I and Fig.~\ref{fig:prompt_comparison}, the standard \texttt{ARCE} achieves the highest Macro-F1 score (77.20\%), outperforming both variants. A deeper analysis reveals a distinct pattern in \texttt{ARCE-role}: it obtains the highest precision (78.73\%) at the cost of the lowest recall (72.88\%). We attribute this to a granularity mismatch. The complex instructions in Strategy B encourage the LLM to generate highly abstract functional descriptions. While intellectually rich, these high-level abstractions do not map linearly to the token-level features required by the RoBERTa encoder, causing the student model to overfit to specific roles while failing to generalize to broader entity mentions.

Furthermore, \texttt{ARCE-think} unexpectedly underperforms the standard \texttt{ARCE}, suggesting that raw CoT traces act as semantic noise during the distillation process. The verbose intermediate reasoning steps dilute the core semantic signal, making it harder for the limited-capacity student model to extract essential entity-label alignments. Collectively, these findings robustly support the less is more principle: for domain adaptation of SLMs, a large volume of simple, explanation-based elucidations provides a cleaner signal-to-noise ratio than complex reasoning chains.

\subsection{Data Efficiency and Scalability}
To strictly evaluate the data efficiency and scalability of our proposed framework, we pre-trained ARCE on monotonically increasing subsets (25\%, 50\%, 75\%, and 100\%) of the generated \texttt{Cote} corpus. As visually illustrated in Figure 3, the model performance exhibits a robust, near-linear positive correlation with the volume of pre-training data. The Macro-F1 score steadily climbs from 74.23\% using only a quarter of the corpus to a peak of 77.20\% with the full dataset.

A critical observation from this analysis is the high value density of our explanation-based corpus. Remarkably, even when utilizing only 25\% of the generated data, ARCE achieves a Macro-F1 score of 74.23\%, which already surpasses the fully supervised RoBERTa-wwm-ext-CRF baseline by over 3 points. This indicates that the contextualized elucidations generated by the LLM provide rich, dense semantic signals that allow the student model to align with the domain logic rapidly, significantly reducing the data volume required for effective adaptation compared to traditional raw-text pre-training.

Furthermore, the performance curve shows no signs of saturation or diminishing returns, which typically plague synthetic data augmentation methods. This sustained growth trajectory suggests that the model's performance boundaries are not yet reached and that further gains are achievable simply by scaling the generation process. It implies that the \texttt{Cote} corpus derives its strength from the cumulative effect of a diverse set of semantic bridges rather than relying on a few golden samples, highlighting the robust scalability of ARCE as a data-centric solution for low-resource domains.

\begin{figure}
\centering
\includegraphics[width=0.45\textwidth]{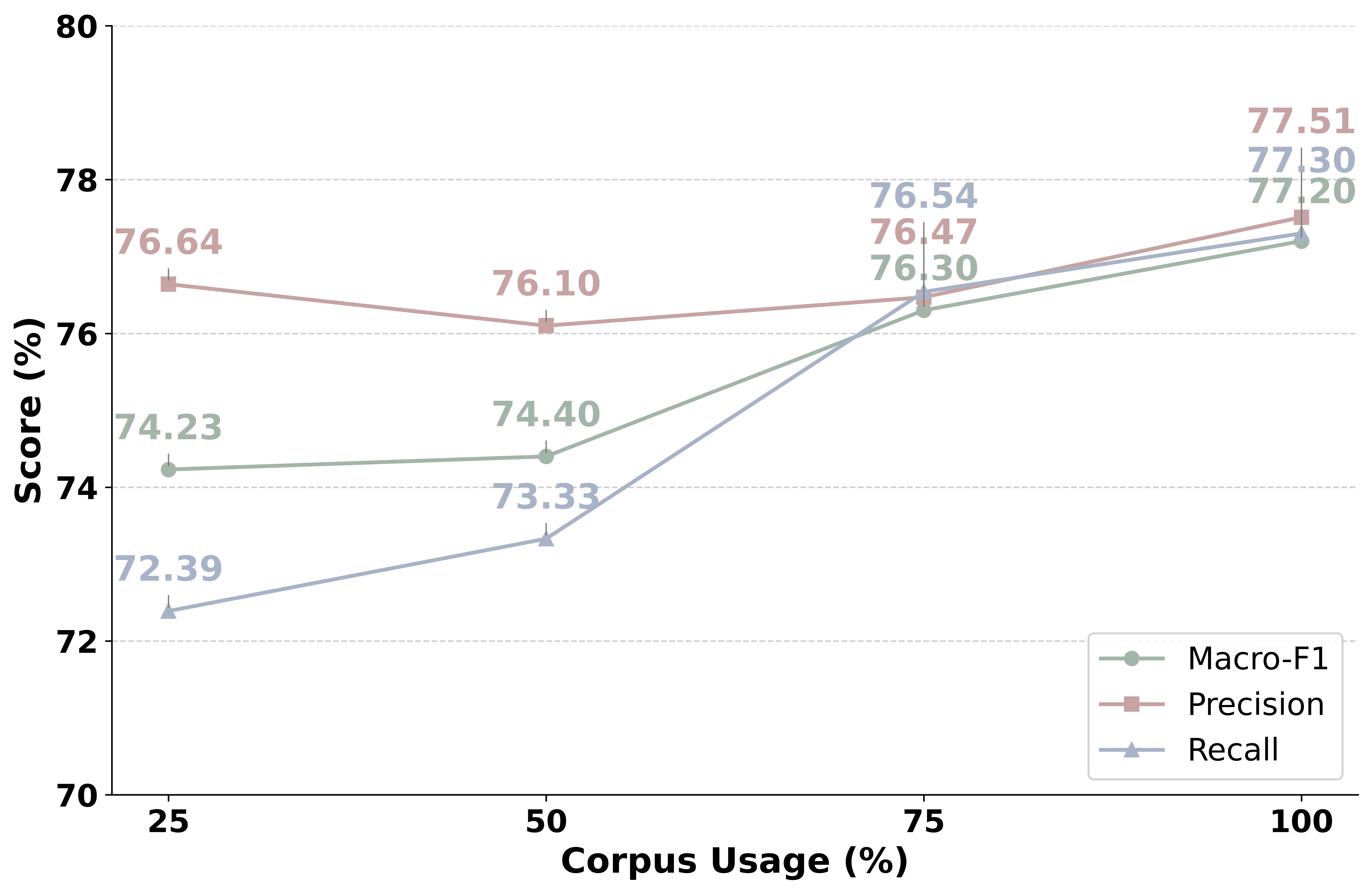}
\caption{Data Efficiency and Scalability Analysis.}
\end{figure}

\section{Conclusion}
In this study, we proposed ARCE, a novel framework that adapts lightweight models to the AEC domain using LLM-generated explanations, establishing a new state-of-the-art with a 77.20\% Macro-F1 score. Our investigation reveals a critical less is more principle: simple, direct elucidations facilitate superior knowledge transfer compared to complex reasoning chains in the NER task, which often introduce semantic noise for smaller models. By achieving high accuracy with minimal parameters, ARCE offers a viable solution for resource-constrained edge deployment. Future work will explore the generalizability of this paradigm across broader domains and its integration into end-to-end automated compliance checking systems.

\section*{Acknowledgment}

The authors would like to thank the anonymous reviewers for their helpful comments and suggestions.

% \balance
% \normalem
\bibliographystyle{unsrt}
\bibliography{ref}

\end{document}